\documentclass[letterpaper]{article} 
\usepackage{aaai25}  
\usepackage{times}  
\usepackage{helvet}  
\usepackage{courier}  
\usepackage[hyphens]{url}  
\usepackage{graphicx} 
\usepackage{natbib}  
\usepackage{caption} 
\usepackage{algorithm}
\usepackage{algorithmic}
\usepackage{multirow}
\usepackage{amsmath}
\usepackage{amssymb}
\usepackage[table]{xcolor}

\usepackage{newfloat}
\usepackage{listings}
\DeclareCaptionStyle{ruled}{labelfont=normalfont,labelsep=colon,strut=off} 
\floatstyle{ruled}
\newfloat{listing}{tb}{lst}{}
\floatname{listing}{Listing}
\title{MSP-MVS: Multi-Granularity Segmentation Prior Guided Multi-View Stereo}

\author{
    Zhenlong Yuan\textsuperscript{\rm 1}, 
    Cong Liu\textsuperscript{\rm 2}, 
    Fei Shen\textsuperscript{\rm 3}, 
    Zhaoxin Li\textsuperscript{\rm 4, \rm 5}\thanks{Corresponding Author.}, 
    Jinguo Luo\textsuperscript{\rm 2}, 
    Tianlu Mao\textsuperscript{\rm 1}, 
    Zhaoqi Wang\textsuperscript{\rm 1}
}
\affiliations{
    \textsuperscript{\rm 1}Institute of Computing Technology, Chinese Academy of Sciences

    \textsuperscript{\rm 2}Harbin Institute of Technology, Shenzhen

    \textsuperscript{\rm 3}Nanjing University of Science and Technology, Nanjing
    
    \textsuperscript{\rm 4}Agricultural Information Institute, Chinese Academy of Agricultural Sciences
    
    \textsuperscript{\rm 5}Key Laboratory of Agricultural Big Data, Ministry of Agriculture and Rural Affairs

    yuanzhenlong21b@ict.ac.cn, liucong@stu.hit.edu.cn, feishen@njust.edu.cn, 

    cszli@hotmail.com, 23s153135@stu.hit.edu.cn, \{ltm, zqwang\}@ict.ac.cn
}

\usepackage{bibentry}

\begin{document}

\maketitle

\begin{abstract}
Recently, patch deformation-based methods have demonstrated significant strength in multi-view stereo by adaptively expanding the reception field of patches to help reconstruct textureless areas. 
However, such methods mainly concentrate on searching for pixels without matching ambiguity (i.e., \textbf{reliable} pixels) when constructing deformed patches, while neglecting the deformation instability caused by unexpected edge-skipping, resulting in potential matching distortions.
Addressing this, we propose MSP-MVS, a method introducing multi-granularity segmentation prior for edge-confined patch deformation. 
Specifically, to avoid unexpected edge-skipping, we first aggregate and further refine multi-granularity depth edges gained from Semantic-SAM as prior to guide patch deformation within depth-continuous (i.e., \textbf{homogeneous}) areas. 
Moreover, to address attention imbalance caused by edge-confined patch deformation, we implement adaptive equidistribution and disassemble-clustering of correlative reliable pixels (i.e., \textbf{anchors}), thereby promoting attention-consistent patch deformation.
Finally, to prevent deformed patches from falling into local-minimum matching costs caused by the fixed sampling pattern, we introduce disparity-sampling synergistic 3D optimization to help identify global-minimum matching costs.
Evaluations on ETH3D and Tanks \& Temples benchmarks prove our method obtains state-of-the-art performance with remarkable generalization. 
\end{abstract}

\section{Introduction}
Multi-View Stereo (MVS) is a technique that employs images captured from different viewpoints to reconstruct 3D scenes or objects. Its diverse application and downstream task spans across object detection \cite{Lu2023, Lu2024}, image denoising \cite{Yang2024c, Li2022b}, video representation \cite{Liu2024, Liu2024a}, multiview clustering \cite{SSGCC, CMSCGC}, etc. Numerous ideas \cite{Chen2024c, Chen2024b, Kang2025} and datasets \cite{Blendedmvs} have advanced its progress. 
However, it still faces the primary challenge of reconstructing large textureless areas.

The majority of traditional MVS methods are extended from the PatchMatch (PM) \cite{PM} algorithm, which predicts depth hypothesis by computing the minimum matching cost between patch pairs and achieves iterative renewal through propagation and refinement. 
However, depth hypotheses may exhibit ambiguity in textureless areas when patches lack reliable pixel correspondence. 
To tackle this, numerous methods attempt to introduce technologies such as cascade architecture \cite{Pyramid}, confidence estimators \cite{CLD-MVS}, superpixel planarization \cite{TSAR-MVS} and planar priors \cite{HPM-MVS}, etc. 
Yet these methods either require complex and time-consuming post-processing or result in excessively smooth surface details. 

Differently, learning-based methods employ deep neural networks to create learnable 3D cost volumes, thereby enabling the extraction of high-dimensional features and thus providing a wider receptive field. 
\begin{figure}
    \centering
    \includegraphics[width=\linewidth]{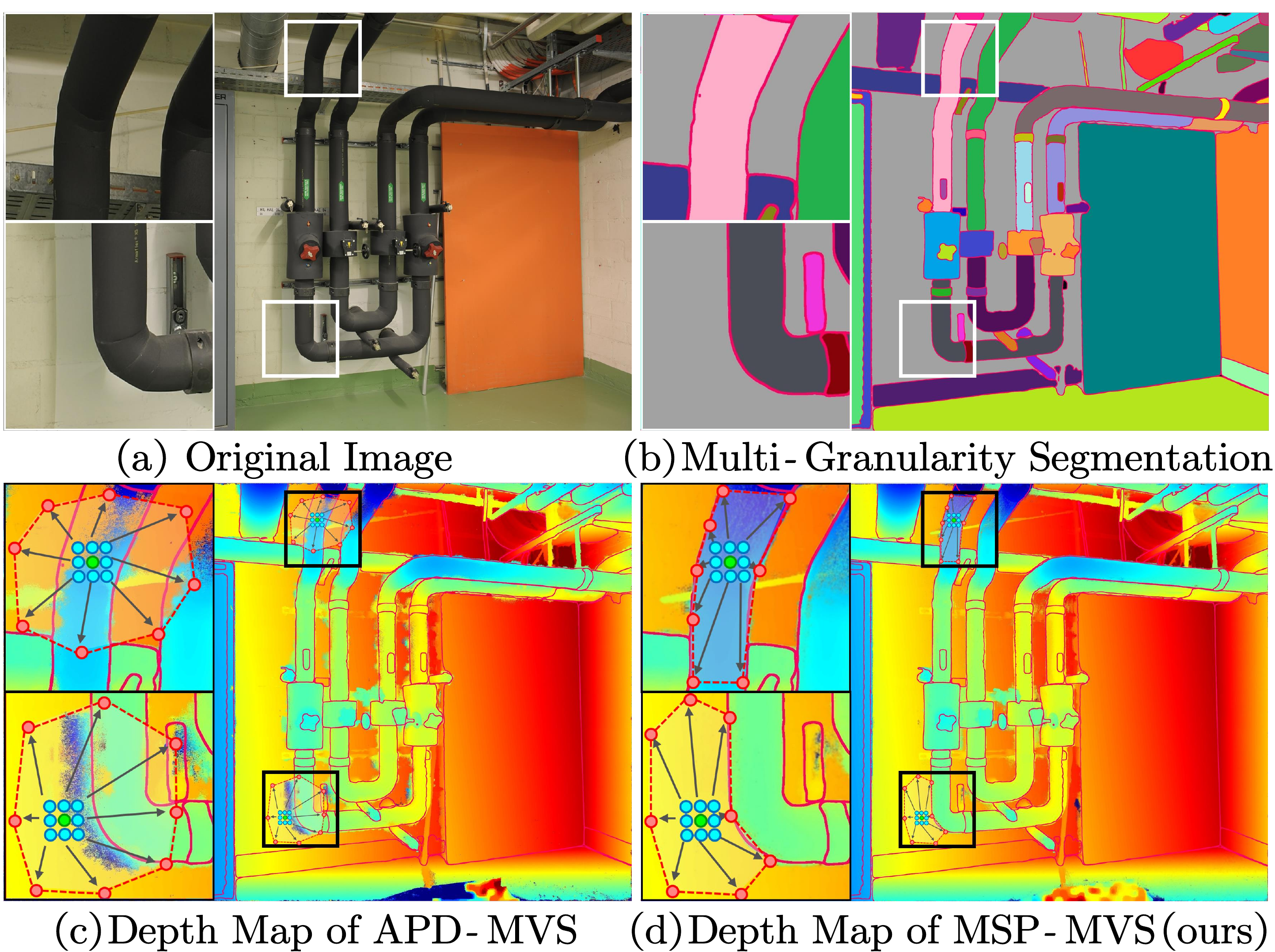}
    \caption{Comparative analysis between APD-MVS and our method. 
    In (c) and (d), green, blue and red dots respectively denote the central pixel, conventional PM and deformable PM. Due to the lack of depth edge guidance in APD-MVS (c), its deformable PM occurs edge-skipping, thereby covering areas with depth-discontinuity. Differently, our method (d) leverages multi-granularity segmentation image (b) as prior to guide deformable PM within homogeneous areas.}
    \label{fig:1}
\end{figure}
Although typical methods employ deformable convolutions \cite{TransMVSNet} or adopt a coarse-to-fine strategy \cite{EP-Net} to broaden the receptive field, their unaffordable time and memory consumption make them unsuitable for high-resolution imagery. 
Addressing this, others attempt to employ lightweight GRU \cite{Hybrid} or leverage vision transformers \cite{MVSFormerPLUS} to compact 3D volumes and enhance prediction quality. 
However, these networks struggle with generalization when faced with scenarios different from their training datasets, posing challenges for practical applications.

Building on advances in traditional and learning-based MVS methods, APD-MVS~\cite{APD-MVS} offers an innovative solution to tackle matching ambiguity. As shown in Fig. \ref{fig:1} (c), centered on the \textbf{unreliable} green pixel (i.e., pixels with matching ambiguity), the algorithm first segments its surrounding area into eight sectors at fixed angles, then selects a reliable red pixel within each sector as \textbf{anchor} pixel to form the deformable PM, which will substitute the blue neighbors in conventional PM to help reconstruct textureless areas. 
The APD-MVS assumes that all areas in deformable PM share the same depths and normals as the central pixel, referred to as \textbf{homogeneous}.
Homogeneous areas are notably characterized by depth continuity, which makes them distinguishable at depth boundaries. 
Nonetheless, in complicated scenes, shadows and occlusions may cause deformable PM erroneously crossing depth edges and consequently being selected into \textbf{heterogeneous} areas (i.e., areas with depth-discontinuity), leading to potential reconstruction failures. 

Therefore, it is crucial to fully exploit depth edges to aid in patch deformation.
However, extracting comprehensive depth edges is challenging.
Previous methods \cite{EP-Net, Xue2019, Bao2014} attempt to extract RGB edges for reconstruction, while the illumination-sensitive and shadow-unaware RGB edges are impractical for complex scenes. Differently, others \cite{Shvets2024, SD-MVS} leverage semantic information for edge extraction, yet single semantic information fails to accurately capture comprehensive depth edges in varying scenes. 
Consequently, a method that effectively integrates semantic information from multiple granularities is urgently needed.


Addressing this, we introduce MSP-MVS, which innovately proposes multi-granularity segmentation prior to facilitate edge-confined patch deformation. Specifically, we first adopt Semantic-SAM \cite{S-SAM} for panoramic segmentation to extract semantic information of images with varying granularities, then we aggregate them to obtain integrated edges that can distinguish different homogeneous areas. 
To better align the integrated edges with actual depth edges, we further apply edge refinement to generate fine-grained depth edges, which are then adopted as the multi-granularity segmentation prior to constrain deformed patches within homogeneous areas, as shown in Fig. \ref{fig:1} (d).

Although edge-confined patch deformation effectively removes deformed patches in heterogeneous areas, it inevitably causes an imbalanced distribution for the remaining patches. Addressing this, we introduce an adaptive equidistribution strategy, which adaptively divides the homogeneous area of each unreliable pixel into eight sectors based on the number of surrounding reliable pixels. Moreover, we further employ a disassemble-clustering strategy to quadruple anchor numbers within each sector and group selected anchors into eight anchor clusters that serve as new deformed patches with attention consistency for matching cost.

While the adaptive equidistribution and disassemble-clustering strategy achieves attention-consistent patch deformation, the fixed sampling pattern within patches may still cause the matching cost to fall into a local-minimum. Therefore, we introduce a disparity-sampling synergistic 3D optimization method, which synergistically and iteratively optimizes sampling pixels and disparities in 3D space to enable deformed patches identify their global minimal matching costs. 
In summary, our contributions are as follows:

\begin{itemize}
    \item We develop MSP-MVS, which leverages Semantic-SAM to aggregate and further refine multi-granularity depth edges as prior for edge-confined patch deformation.
    \item We propose adaptive equidistribution for sector division and disassemble-clustering for anchor clustering to facilitate attention-consistent patch deformation.
    \item We present the disparity-sampling synergistic 3D optimization which optimizes both sampling pixels and disparities to help identify global-minimum matching costs.
\end{itemize}

\section{Related Work}
\subsection{Traditional Patchmatch-based Methods}
The fundamental concept of PatchMatch \cite{PM} employs random initialization, propagation and refinement to match optimal patch pairs between images. 
PMS \cite{PMS} pioneers in introducing this in stereo vision, facilitating the emergence of numerous future works.
For acceleration, Gipuma \cite{Galliani} proposes checkerboard diffusion propagation to achieve algorithm deployment on GPUs.
To reconstruct textureless areas, ACMM \cite{ACMM} presents both multi-scale architecture and multi-view consistency, improved by ACMMP \cite{ACMMP} which introduces triangularization plane priors for post-processing.
Differently, TAPA-MVS \cite{TAPA-MVS}, PCF-MVS \cite{PCF-MVS} and TSAR-MVS \cite{TSAR-MVS} attempt to employ superpixel planarization to reconstruct textureless areas, while their performance is severely affected by limited superpixel size.
Moreover, HPM-MVS \cite{HPM-MVS} introduces non-local extensible sampling patterns to avoid local optimal solutions, but its performance is limited by fixed patch receptive fields. Addressing this, SD-MVS \cite{SD-MVS} employs instance segmentation for patch deformation within instances. However, its results severely depend on low-precision segmentation results. Moreover, APD-MVS \cite{APD-MVS} separates patches of unreliable pixels into several outward-spreading anchors with high reliability, while the absence of edge constraint may cause unexpected edge-skipping and deformation instability.

\begin{figure*}
\centering
\includegraphics[width=\linewidth]{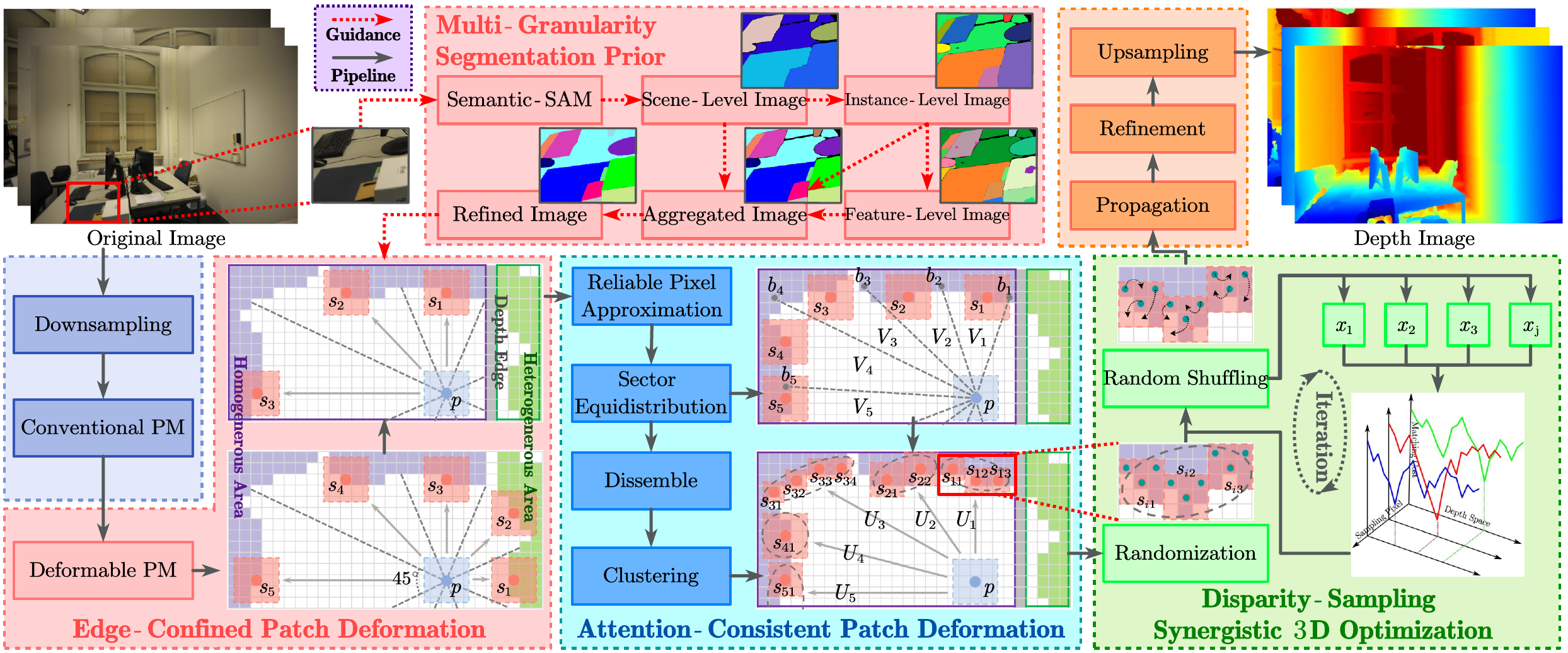}
\caption{
Pipeline of MSP-MVS. We first adopt Semantic-SAM to obtain multi-granularity segmentation images. We then aggregate and further refine these images as multi-granularity segmentation prior to facilitate edge-confined patch deformation. Subsequently, we propose adaptive equidistribution for sector division and disassemble-clustering strategy for anchor clustering to promote attention-consistent patch deformation. Additionally, we introduce disparity-sampling synergistic 3D optimization to help deformed patches identify their global-minimal matching costs. After several iterations we obtain final depth images.
}
\label{fig:pipeline}
\end{figure*}

\subsection{Learning-based MVS Methods}
Recently, the advancement of deep learning has led to the emergence of numerous large-language models \cite{Luo2024, Luo2024a}, gaussian splatting algorithms \cite{Gan2023a, Gan2023}, and self-supervised learning methods \cite{Li2022a, Song2025, chen2023self, chen2024learning, SUN1, SUN2} and vision-language model \cite{chen2024bimcv, qianmaskfactory, chen2024tokenunify}.
In MVS field, MVSNet \cite{MVSNet} is the first to utilize differentiable 3D cost volumes to construct deep neural networks.
To reduce memory, Cas-MVSNet \cite{Cas-MVSNet} adopts a coarse-to-fine strategy to retain depths across multiple scales.
IterMVS-LS \cite{IterMVS} leverages a lightweight probability estimator to encode depth distribution during regularization.
For feature extraction, AAR-MVSNet \cite{Aa-Rmvsnet} employed deformable convolutions to achieve adaptive feature aggregation.
Differently, MVSTER \cite{MVSTER} leverages the transformer structure to introduce the multi-head attention mechanism.
Geo-MVSNet \cite{GeoMVSNet} adopts a two-branch geometry fusion network to enhance geometry perception. 
RA-MVSNet \cite{RA-MVSNet} associates hypothesis planes with surface patches to enhance perception field. 
Moreover, Raymvsnet++ \cite{RayMVSNet} applies an epipolar transformer to enhance feature aggregation along camera rays. GoMVS \cite{GoMVS} integrates a geometrically consistent propagation module to refine cost aggregation. Differently, our method targets on aggregating multi-granularity depth edges for edge-confined patch deformation.
Despite this, most of them still suffer from either unaffordable memory costs or limited generalization. 

\section{Preliminaries}
The conventional PM method which utilizes fixed patch for reconstruction should be detailed at first.
Given a series of input images $\mathcal{I} = \{I_n \mid n=1 \cdots N\}$ and corresponding camera parameters $\mathcal{P}=\{K_n, R_n, C_n \mid n=1 \cdots N\}$, each image is selected from $\mathcal{I}$ to be the reference image $I_{i}$,  with its depth map sequentially reconstructed through pairwise matching with the remaining images $(\mathcal{I} - I_{i})$.  
Here $K$, $R$ and $C$ respectively represent the intrinsic parameters, the rotation matrix and the camera center.
Specifically, given reference image $I_{i}$ and source image $I_{j}$, the homography matrix $H_{ij}$ for pixel $p$ in $I_i$ is defined as follows:
\begin{equation}
H_{i j}=K_j\left(R_j R_i^{-1}+\frac{R_j\left(C_i-C_j\right) n^T}{n^T d K_i^{-1} p}\right) K_i^{-1},
\end{equation}
where $n^T$ and $d$ respectively denote normal and depth hypothesis. Given the fixed patch which is a square window $W$ centered on $p$ in $I_{i}$, we can leverage $H_{ij}$ to obtain its corresponding patch $W'$ in $I_{j}$ . Then the matching cost is:
\begin{equation}
m_{i, j}\left(p, W\right)=1-\frac{cov\left(W, W'\right)}{\sqrt{cov\left(W, W\right) cov\left(W', W'\right)}},
\end{equation}
where $cov$ is weighted covariance \cite{COLMAP}. 
Finally, we can obtain multi-view aggregated cost via view weights \cite{ACMM} and further leverage propagation and refinement to update depth with the minimum cost.

Differently, the deformable PM method \cite{APD-MVS} calculates costs by constructing multiple patches centered on different anchors to form the deformed patches. 
Anchors are reliable pixels identified by a spoke-like searching strategy centered on unreliable pixel $p$ with $\theta$ as the radius. Then the deformable matching cost is calculated as:
\begin{equation}
m_{ij}(p, S) = \lambda m(p, W) + (1 - \lambda) \frac{\sum_{s \in S} m(s, W_s)}{|S|},
\end{equation}
where $S$ is a collection of anchors. Both $W$ and $W_s$ share the same window size $w \times w$ while respectively have intervals $\frac{w}{2}$ and 2. In experiment, $w = 11$, $\lambda = 0.25, \theta = 45^{\circ}, |S| = 8$.

\begin{figure*}
\centering
\includegraphics[width=\linewidth]{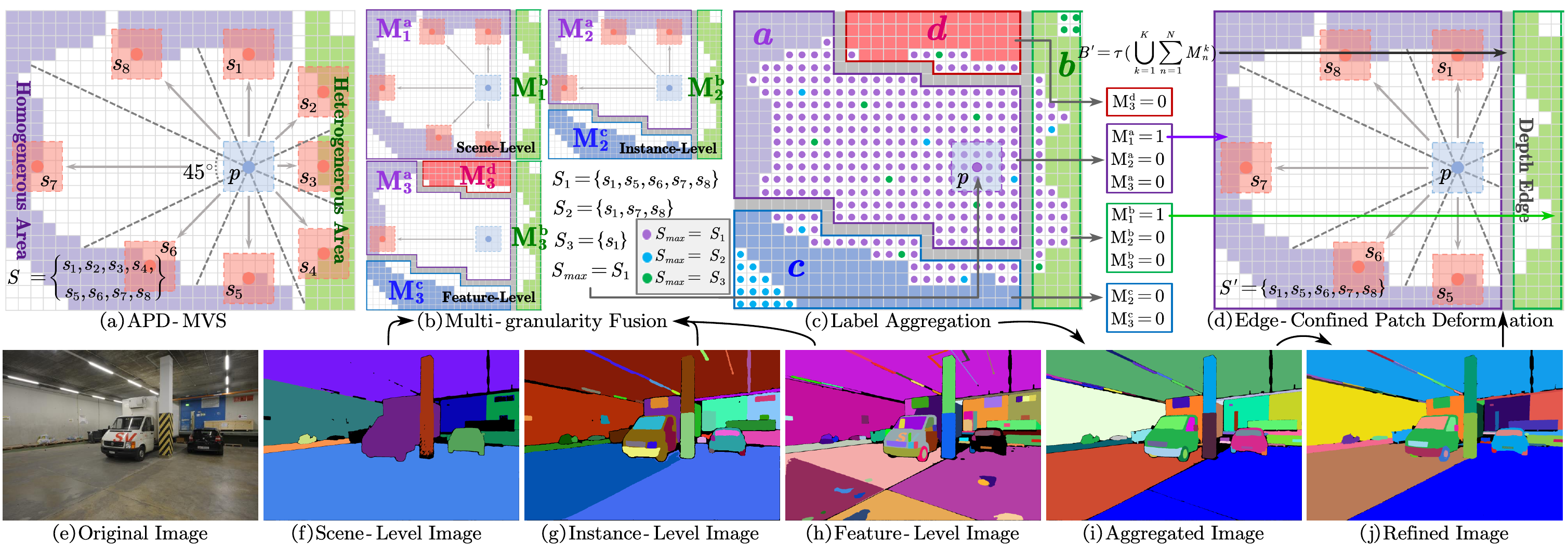}
\caption{Multi-Granularity Segmentation Prior. From (a) to (d), purple, green and gray backgrounds respectively denote homogeneous areas, heterogeneous areas and depth edges, with sectors divided at fixed angle $45^{\circ}$ by black dash lines. In (c), purple, blue, and green dots respectively denote pixels whose optimal anchor subsets $S_{max}$ equals $S_1$, $S_2$, and $S_3$. Since most pixels are purple in (c) (i.e.,  $S_{max} = S_1$), the scene-level masks $M^a_1$ and $M^b_1$ in (b) are reliable, while others are misidentified masks. 
}
\label{fig: SAM}
\end{figure*}

\section{Method}
The overview of the proposed method is illustrated in Fig. \ref{fig:pipeline}, and each component will be detailed in subsequent sections.

\subsection{Multi-granularity Segmentation Prior}
As shown in Fig. \ref{fig: SAM} (a), since APD-MVS lacks depth edge guidance, its anchors typically skip depth edges to be selected within green heterogeneous areas, causing potential inaccuracy.
Therefore, we attempt to extract depth edges as guidance for patch deformation.
However, without accurate depth maps, it is rather difficult to extract depth edges. Although depth edges and image edges have a certain degree of co-occurrence, they are easily confused by factors such as illumination, shadow, occlusion, etc.
Differently, by introducing semantic awareness and granularity richness, Semantic-SAM can achieve panoramic segmentation at any granularity level, which may contain potential depth edges.
Therefore, we attempt to adopt Semantic-SAM to extract multi-granularity depth edges for patch deformation. 

\subsubsection{Edge Aggregation. }
As shown in Fig. \ref{fig: SAM}, we first employ Semantic-SAM for panoramic segmentation on input image $I_i$ in (e) to generate mask maps in (f), (g) and (h) with varying granularities from coarse to fine (i.e., scene, instance and feature level). Then we have $\{M_n | n = 1, 2, 3\}$, where $M_n$ denotes the $n^{th}$ mask map. In addition, we further apply a transformation function $\tau$ to identify the boundaries of different mask maps $M_n$. Mathematically, $B_n=\tau\left(M_n\right)=\left\{(x, y) \mid \Delta M_n(x, y) \neq 0\right\},$, where $\Delta$ represents the Laplace operator. Then we have $\{B_n = \tau(M_n) | n = 1, 2, 3\}$, where $B_n$ denotes the $n^{th}$ boundary map. 
Then for each unreliable pixel $p$ in $I_i$, we first construct an anchor subset $S_n$ for the $n^{th}$ boundary map $B_n$. 
Subsequently, we utilize the nth boundary map $B_n$ to determine whether each anchor $s_i \in S$ is retained. Specifically, an anchor $s_i$ can only be retained if its line $l$, connecting it with the center pixel $p$, does not intersect with $B_n$; otherwise, we discard it. As shown in Fig. \ref{fig: SAM} (b), All retained $s_i$ form $S_n$, defined by: $\{s_i \in S_n \mid l \cap B_n=\emptyset \}$.
Therefore, we have $\{M_n, B_n, S_n | n = 1, 2, 3\}$.

We then attempt to obtain the optimal anchor subset $S_{max}$ based on the reliability of anchor distributions for different anchor subsets.
Specifically, we perform RANSAC-based planarization on all anchors in $S_n$ to obtain the number of anchors classified as inliers, termed as $A_n$. 
Then we consider that $S_{max}$ should maximize the number of inliers $A_n$ and maximize the proportion of inliers $\frac{A_n}{|S_n|}$, defined by:
\begin{equation}
S_{max} = \arg \max_{S_n} \left( \max \left(A_n \right) \text{ s.t. } \max \left(\frac{A_n}{|S_n|}\right) \right).
\end{equation}
However, directly replacing $S$ with $S_{max}$ could render deformed patches susceptible to noise disturbances or local-minimum. Instead, a more robust approach involves merging different mask maps $M_n$, thereby forming an aggregated segmentation image that offers enhanced guidance. Therefore, we define $M_n^k$ as the $k^{th}$ label in the $n^{th}$ layer mask map. 
Subsequently, as depicted on the right side of Fig. \ref{fig: SAM} (c), we retain $M_n^k$ only if $n^{th}$ layer contains the highest number of optimal anchor subsets among all layers, formulated by:
\begin{equation}
M_n^k = \begin{cases} 
1, & \text { if } n=\arg\max_l \left( \sum_{p \in M_l^k} \mathbb{I}\left(S_{max} = S_l\right) \right); \\
0, & \text { else. }
\end{cases}
\end{equation}
where $\mathbb{I}(\cdot)$ is the indicator function such that $\mathbb{I}(\text {true}) = 1$ and $\mathbb{I}(\text {false}) = 0$. Therefore, reliable masks that maximize the quantity of $S_{max}$ are effectively retained.
Finally, as shown in Fig. \ref{fig: SAM} (i), by aggregating all label masks among all layers, we effectively obtain the aggregated image, which is then further used by function $\tau$ to derive the aggregated boundary map:
$B^{\prime} = \tau(\bigcup_{k=1}^{K} \sum_{n=1}^{N}  M_n^k)$.
$B^{\prime}$ is then adopted as the multi-granularity segmentation prior to constrain each unreliable pixel $p$ in $I_i$ to obtain the new anchor collection $S^{\prime}$ for patch deformation as shown in Fig. \ref{fig: SAM} (d). Although $S^{\prime}$ contains fewer anchors than $S$, it significantly enhances the anchor's reliability.
Compared with APD-MVS in Fig. \ref{fig: SAM} (a), our edge-confined patch deformation in Fig. \ref{fig: SAM} (d) effectively avoids unexpected edge-skipping and constraints patch deformation in homogeneous areas, thus avoiding distortion.


\subsubsection{Edge Refinement. }
Moreover, to better align the integrated edges with their actual depth edges, we further propose an edge refinement strategy leveraging the conditional random field (CRF) algorithm.
CRF is composed of both unary potential and pairwise potential. The unary potential $\psi_u\left(l_i\right)$ represents the probability of the pixel $p$ being assigned to the label $l_i$. 
The pairwise potential $\psi_p\left(l_i, l_j\right)$ describes the joint probability of two adjacent pixels $p$ and $q$ being categorized as label $l_i$ and $l_j$, respectively. 
Since edges primarily exhibit depth and color differences, we incorporate both geometry and color similarities for edge refinement, such that:
\begin{equation}
\psi_p\left(l_i, l_j\right)= \mu\left(l_i, l_j\right) \cdot e^{-(t \cdot \frac{\left\|D_p-D_q\right\|}{\alpha} + \frac{1}{t} \cdot \frac{\left\|I_p \cdot I_q - 1\right\|}{\beta})},
\end{equation}
where $D_p$, $D_q$, $I_p$ and $I_q$ respectively denote the depth and normalized RGB colors of pixels $p$ and $q$, $t$ is the current iteration number, $\mu\left(l_i, l_j\right)$ is the label compatibility function defined in \cite{CRF}. 
Finally, as shown in Fig. \ref{fig: SAM} (j), by minimizing both $\psi_u\left(l_i\right)$ and $\psi_p\left(l_i, l_j\right)$ among all pixels, we identify the optimal label assignment, thereby generating fine-grained depth edges for accurate patch deformation. A detailed description is available in supplementary material. 



\subsection{Attention-Consistent Patch Deformation}
Although edge-confined patch deformation in Fig. \ref{fig: Sector Averaging} (b) effectively improves the reliability of deformed patches compared to APD-MVS in Fig. \ref{fig: Sector Averaging} (a), it may cause the loss of anchors within sectors fully enclosed by depth edges.
Addressing this, we propose adaptive equidistribution, which uniformly divides the homogeneous area of each unreliable pixel into eight sectors based on the number of its surrounding reliable pixels to recover previously-discarded anchors.

\subsubsection{Adaptive Equidistribution. }
\begin{figure}
\centering
\includegraphics[width=\linewidth]{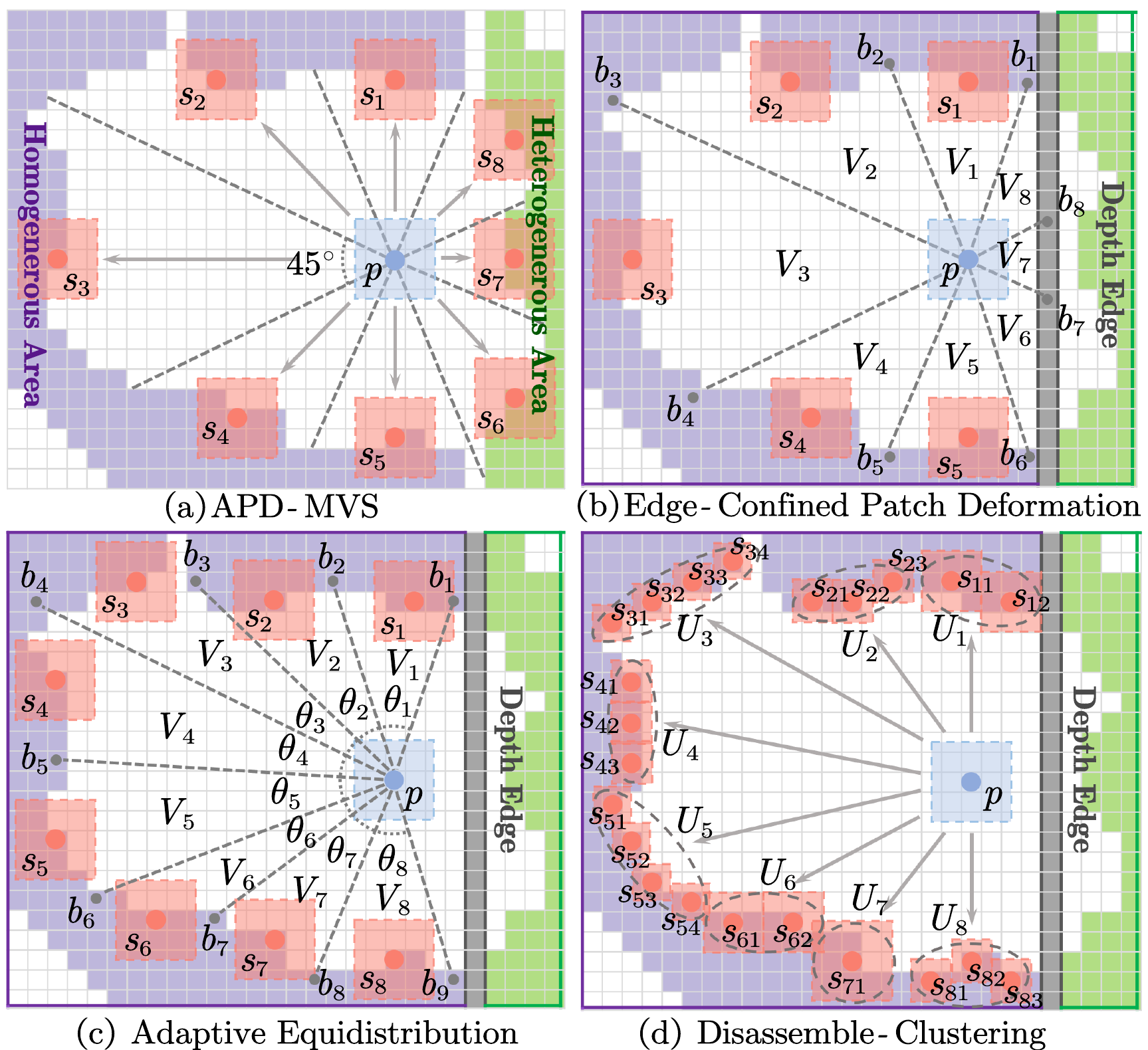}
\caption{Attention-Consistent Patch Deformation. Blue and red dots respectively denote the central pixel and anchors. purple, green and gray backgrounds respectively denote homogeneous areas, heterogeneous areas and depth edges.
}
\label{fig: Sector Averaging}
\end{figure}

Specifically, as shown in Fig. \ref{fig: Sector Averaging} (b), centered on unreliable pixel $p$, we first emit 8 rays at fixed angle $45^{\circ}$ for sector division. We then define the first reliable pixel or depth edge encountered by each ray as the boundary pixel.
Thus, we have all sector angles $\Theta=\{\theta_i\mid i=1 \cdots 8\}$, sectors $\mathcal{V}=\{V_i \mid i=1 \cdots 8\}$, and boundary pixels $\mathcal{B}=\{b_i, b_{i+1}\mid i=1 \cdots 8\}$. Subsequently, for sector $V_i$, if either its boundary pixels $b_i$ or $b_{i+1}$ is located on depth edges rather than reliable pixels, we consider $V_i$ to be fully enclosed by depth edges and contain no reliable pixels. Otherwise, we consider $V_i$ to be fully enclosed by a certain number of reliable pixels, denoted as $n_i$.
Since reliable pixels are typically continuously distributed, $n_i$ can be approximately estimated as the L1 distance between two boundary pixels $b_i$ and $b_{i+1}$.
Moreover, we can obtain the average number of all reliable pixels $\bar{d}$ among all sectors $\mathcal{V}$. Then through the weighted averaging of all reliable pixels, we can proportionally recalculate all angles $\Theta'=\{\theta_i'\mid i=1 \cdots 8\}$ and in turn acquire new sectors $\mathcal{V}'$ and boundary pixels $\mathcal{P}'$ for allocation. The $k^{th}$ rectified angle $\theta_k'$ is calculated by:
\begin{equation} 
\theta_k^{\prime}=\theta_{k-1}^{\prime}+\frac{k \bar{d}-\sum_{i=1}^{x-1} d_i}{d_{x}}, \sum_{i=1}^{x-1} d_i \leqslant k \bar{d} < \sum_{i=1}^{x} d_i.
\end{equation}
Iterating the above process several times we obtain Fig. \ref{fig: Sector Averaging} (c), where each sector is assigned a roughly equal number of reliable pixels to ensure a balanced division without wastage.

\subsubsection{Disassemble-Clustering. }
After adaptive equidistribution, all anchors including the removed ones are evenly allocated.
However, since each sector only selects one anchor to construct the deformed patch, a single patch typically fails to comprehensively capture the characteristics of all reliable pixels within its sector, thus causing insufficient attention.

Addressing this issue, we first quadruple the number of anchors within each sector to disassemble each anchor-centered patch into four smaller sub-patches. 
We then further perform clustering on all anchors to obtain eight anchor clusters that served as the new deformed patches with attention consistency. 
Specifically, we apply RANSAC-based planarization on all quadrupled anchors within $\mathcal{V}'$ to derive the optimal plane $\pi$. Subsequently, we take $\varepsilon$ as a threshold to filter out outliers, selecting anchors whose distance to $\pi$ is less than $\varepsilon$ as reliable anchors for subsequently clustering. 

Here we adopt the density-based DBSCAN algorithm for clustering, which is robust against noise and more suitable for randomly distributed anchors compared to K-means. 
Specifically, anchors with $\eta$ neighbors within radius $\gamma$ are classified as core points. These core points are then connected if they are within $\gamma$ of each other to form different clusters. Other unconnected points are either integrated into existing clusters if they are within $\gamma$ of any connected core points, or form an independent cluster if isolated from each other. The above process will iteratively repeat to obtain $N$ anchor clusters, defined by $\mathcal{U}=\{U_i \mid i=1, \ldots, N\}$, each comprising $k_i$ anchor that $\mathcal{K}=\{k_i \mid i=1, \ldots, N\}$, as shown in Fig. \ref{fig: Sector Averaging} (d). For cluster $U_i$, all its anchors are $S_i'=\{s_{ij} \mid j=1, \ldots, k_i\}$, with their patch size set to $\frac{w}{k_i}$ to balance attention, thereby avoiding patches overlapping with each other in dense areas.
Ultimately, the aggregation of all anchors within all clusters $S^{\prime\prime} = \sum_{i=1}^{N} S_i'$ will replaces $S^{\prime}$ for attention-consistent patch deformation. Comparing with Fig. \ref{fig: Sector Averaging} (b), the sub-patches in Fig. \ref{fig: Sector Averaging} (d) provides more balanced coverage of reliable pixels in homogeneous areas.

\subsection{Disparity-Sampling Synergistic 3D Optimization}
Fig. \ref{fig:candidate} (a) depicts a close-up view of pixel $p$ selecting one of its anchor pixels $s_i$ for patch deformation. Centered on $s_i$, a sub-patch is constructed and fixed-interval sampling pixels are adopted for cost computation. 
Subsequently, as shown in the cyan line of Fig. \ref{fig:candidate} (b), by assigning varying disparities to these sampling pixels,
APD-MVS constructs a cost profile in 2D space to help identify 'global' minimum matching costs.

However, fixed-interval sampling pixels typically fail to globally represent the features of its sub-patch. 
Therefore, its cost profile is completely different from the optimal cost profile (i.e. the red line of Fig. \ref{fig:candidate} (b)) which reflects the real matching cost, thereby causing costs to be trapped in a local minimum rather than the global minimum.
Addressing this, we attempt to synergistically optimize sampling pixels and disparities to construct multiple cost profiles in 3D space, thereby identifying the real global-minimum matching cost.

Considering identifying optimal sampling pixels in sub-patches is a discrete optimization problem, we employ iterative local search for optimization, which is achieved by performing random perturbation on current solutions and minimizing an objective function to identify the global optimum.


\subsubsection{Sampling Pixel Randomization. }
\begin{figure}
\centering
\includegraphics[width=\linewidth]{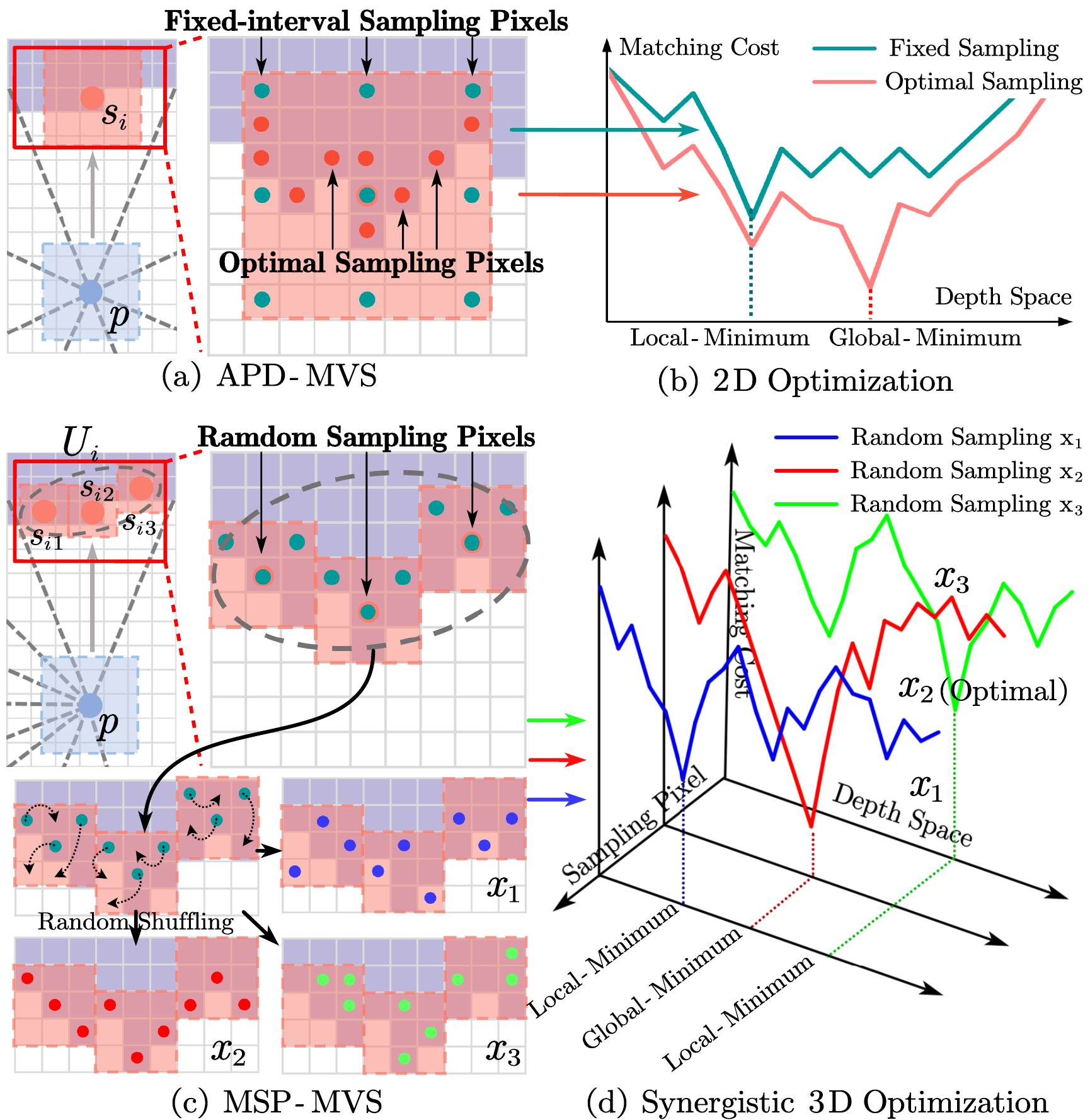}
\caption{Disparity-Sampling Synergistic 3D Optimization. (a) and (b) respectively represent the fixed sampling pattern and 2D cost optimization employed by APD-MVS.
(c) and (d) respectively illustrate the sampling pixel randomization and 3D cost profile optimization of our proposed method. }
\label{fig:candidate}
\end{figure}

Specifically, as shown in Fig. \ref{fig:candidate} (c), to maintain the total number of sampling pixels within deformed patches, we first assign each anchor-centered sub-patch in cluster $U_i$ a certain number of sampling pixels, calculated as $ \frac{9}{k_i}$. These sampling pixels are initially positioned randomly for cost computation, which we refer to as $x_0$. Next, we generate multiple solutions by randomly shuffling their positions to acquire a solution set $\mathcal{X}=\{x_j \mid j=1, \ldots, J\}$.
For each solution $x_j$, given pixel $p$ and current disparity $d$, we construct a disparity sequence $D$ centered on $d$ with length $\mu$ and interval $\delta$, defined as:
\begin{equation}
D = \left\{ d_i \mid d_i = d + \left(i - \frac{\mu + 1}{2}\right) \cdot \delta, , i=1,2, \ldots, \mu \right\}.
\end{equation}
Subsequently, for each solution $x_j$, we apply each disparity $d_i$ in the sequence $D$ on $x_j$ for cost computation, thereby yielding a cost profile $C_j=\{c_i \mid i=1, \ldots, \mu\}$ and corresponding average cost $\bar{c_j}$. 
By aggregating all solutions, we have $\{(x_j, C_j, \bar{c_j}) \mid j=1, \ldots, J\}$.
Since we introduce multiple solutions $x_j$ to construct distinct 2D cost profiles $C_j$, this can be considered to be a 3D optimization problem. 

\subsubsection{3D Cost Profile Optimization. }
By comparing cost profiles of different solutions, we can identify the best sampling solution for cost computation, as illustrated in Fig. \ref{fig:candidate} (d). The optimal cost profile is defined as the one that minimizes cost while maximizing variance. The minimum cost indicates the best matching results, whereas the maximum variance signifies the lowest matching ambiguity. Specifically, the objective function for the cost profile $C_j$ of solution $x_j$ is:
\begin{equation}
F(x_j) = \sum_{i=1}^{\mu} c_i +  \omega \sum_{i=1}^{\mu} (c_i - \bar{c})^{-2}.
\end{equation}
Finally, by evaluating $\min_{x_j \in \mathcal{X}} F\left(x_j\right)$ we determine the optimal solution $x_j$ among all $\mathcal{X}$, which will replace the initial solution $x_0$ in subsequent optimization. Through multiple iterations, we can progressively refine the sampling pixel positions and approach the global-minimum matching cost.

\begin{figure*}
\centering
\includegraphics[width=\linewidth]{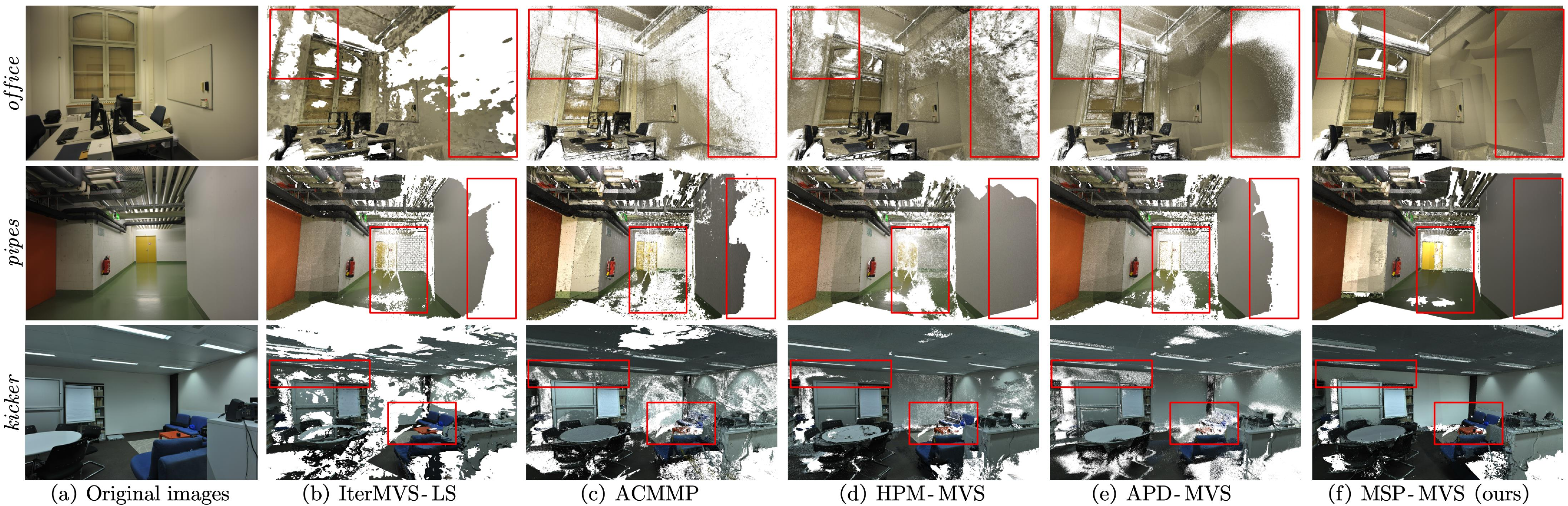}
\caption{An illustration of qualitative results on partial scenes of ETH3D datasets (\emph{office}, \emph{pipes} and \emph{kicker}). Partial challeng areas are shown in red boxes. Our method can effectively reconstruct large textureless areas without introducing detail distortion.
}
\label{fig: eth3d results}
\end{figure*}

\section{Experiment}
\subsection{Datasets and Implementation Details}
We evaluate our work on the ETH3D benchmark \cite{ETH3D} and the Tanks \& Temples benchmark (TNT) \cite{TNT}. 
We compare our method against recent learning-based methods like PatchMatchNet, IterMVS, MVSTER, AA-RMVSNet, EPP-MVSNet, EPNet and traditional methods like TAPA-MVS, PCF-MVS, ACMM, ACMP, ACMMP, SD-MVS, APD-MVS and HPM-MVS.

Our method is implemented on a machine with an Intel(R) Xeon(R) Silver 4210 CPU and eight NVIDIA GeForce RTX 3090 GPUs. Experiments are performed on original images in both ETH3D dataset and TNT dataset.
In cost calculation, we adopt the matching strategy of every other row and column. 
We take APD-MVS \cite{APD-MVS} as our baseline. 


\begin{table}
  \centering
  \renewcommand{\arraystretch}{1.1} 
    \resizebox{\linewidth}{!}{
        \begin{tabular}{c|ccc|ccc}
        \hline
        \multirow{2}{*}{Method} & \multicolumn{3}{c|}{Train} & \multicolumn{3}{c}{Test} \\
        \cline{2-7} & F$_1$ & Comp. & Acc. & F$_1$ & Comp. & Acc. \\   
        \hline 
        \multirow{1}{*}{PatchMatchNet} & 64.21 & 65.43 & 64.81 & 73.12 & 77.46 & 69.71 \\  
        \multirow{1}{*}{IterMVS-LS} & 71.69 & 66.08 & 79.79 & 80.06 & 76.49 & 84.73 \\  
        \multirow{1}{*}{MVSTER} & 72.06 & 76.92 & 68.08 & 79.01 & 82.47 & 77.09 \\  
        \multirow{1}{*}{EPP-MVSNet} & 74.00 & 67.58 & 82.76 & 83.40 & 81.79 & 85.47 \\  
        \multirow{1}{*}{EPNet} & 79.08 & 79.28 & 79.36 & 83.72 & 87.84 & 80.37 \\  
        \hline 
        \multirow{1}{*}{TAPA-MVS} & 77.69 & 71.45 & 85.88 & 79.15 & 74.94 & 85.71 \\  
        \multirow{1}{*}{PCF-MVS} & 79.42 & 75.73 & 84.11 & 80.38 & 79.29 & 82.15 \\  
        \multirow{1}{*}{ACMM} & 78.86 & 70.42 & \textbf{90.67} & 80.78 & 74.34 & \underline{90.65} \\  
        \multirow{1}{*}{ACMMP} & 83.42 & 77.61 & \underline{90.63} & 85.89 & 81.49 & \textbf{91.91} \\  
        \multirow{1}{*}{SD-MVS} & 86.94 & 84.52 & 89.63 & 88.06 & 87.49 & 88.96 \\  
        \multirow{1}{*}{HPM-MVS++} & \underline{87.09} & \underline{85.64} & 88.74 & \underline{89.02} & \underline{89.37} & 88.93 \\  
        \multirow{1}{*}{APD-MVS (base)} & 86.84 & 84.83 & 89.14 & 87.44 & 85.93 & 89.54 \\  
        \hline 
        \multirow{1}{*}{MSP-MVS (ours)} & \textbf{88.09} & \textbf{87.87} & 88.51 & \textbf{89.51} & \textbf{90.38} & 89.08 \\  
        \hline
        \end{tabular}%
    }
  \caption{Quantitative results on ETH3D dataset at threshold $2cm$. Our method achieves the highest F$_1$ and completeness.}
  \label{table: eth3d results}%
\end{table}%

\subsection{Results on ETH3D and TNT}
Qualitative results on ETH3D are illustrated in Fig. \ref{fig: eth3d results}. It is evident that our method delivers the most complete and realistic reconstructed point clouds, especially in large textureless areas like the floors and walls, without introducing detail distortions.
More qualitative results on the ETH3D and TNT dataset can be referred to in the supplementary material. 

Quantitative results on the ETH3D and the TNT datasets are respectively presented in Tab. \ref{table: eth3d results} and Tab. \ref{table: TNT results}. Note that the first group is learning-based methods and the second is traditional methods. Meanwhile, we mark the best results in bold and underline the second-best results.
Our method achieves the highest F$_1$ score and completeness on the ETH3D dataset and the TNT Intermediate dataset, validating its state-of-the-art performance and strong generalization capability. 
Meanwhile, our method achieves the second best results in the TNT advanced datasets, falling short by $0.26\%$ in F$_1$ score compared to EPNet \cite{EP-Net}.

\begin{figure*}
\centering
\includegraphics[width=\linewidth]{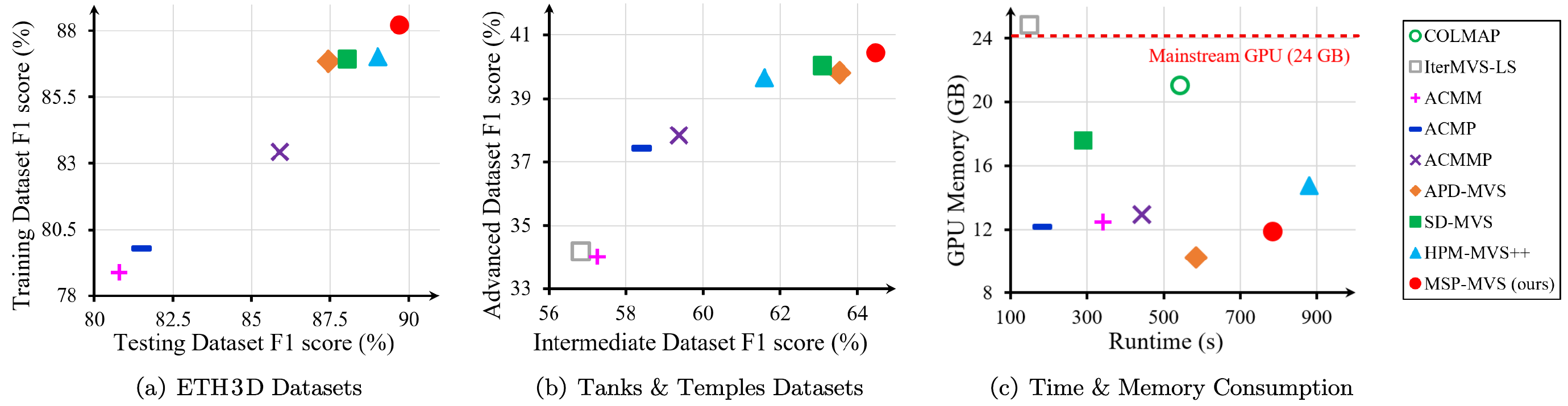}
\caption{Comparative analysis. (a) and (b) respectively illustrate F$_1$ scores among different methods on ETH3D and Tanks \& Temples datasets. (c) depicts their corresponding GPU memory usage (GB) and runtime (second) on ETH3D training datasets.}
\label{fig:mem and time}
\end{figure*}

\subsection{Memory \& Runtime Comparison}
To visually demonstrate the reconstruction quality between different methods, we present the quantitative results of our approach against other methods on both the ETH3D and Tanks \& Temples (TNT) datasets, as shown in Fig. \ref{fig:mem and time} (a) and Fig. \ref{fig:mem and time} (b), respectively.
As illustrated, our method achieves the highest F1 scores on both the ETH3D and TNT datasets, outperforming other state-of-the-art methods such as ACMMP, APD-MVS, SD-MVS, and HPM-MVS, thereby demonstrating its superior performance and robustness.

\begin{table}
    \centering
    \renewcommand{\arraystretch}{1.1} 
    \resizebox{\linewidth}{!}{
        \begin{tabular}{c|ccc|ccc}
        \hline
        \multirow{2}{*}{Method} & \multicolumn{3}{c|}{Intermediate} & \multicolumn{3}{c}{Advanced} \\
        \cline{2-7} & F$_1$ & Rec. & Pre. & F$_1$ & Rec. & Pre. \\   
        \hline 
        \multirow{1}{*}{PatchMatchNet} & 53.15 & 69.37 & 43.64 & 32.31 & 41.66 & 27.27 \\  
        \multirow{1}{*}{AA-RMVSNet} & 61.51 & 75.69 & 52.68 & 33.53 & 33.01 & 37.46 \\  
        \multirow{1}{*}{IterMVS-LS} & 56.94 & 74.69 & 47.53 & 34.17 & 44.19 & 28.70 \\  
        \multirow{1}{*}{MVSTER} & 60.92 & \underline{77.50} & 50.17 & 37.53 & 45.90 & 33.23 \\  
        \multirow{1}{*}{EPP-MVSNet} & 61.68 & 75.58 & 53.09 & 35.72 & 34.63 & \underline{40.09} \\
        \multirow{1}{*}{EPNet} & \underline{63.68} & 72.57 & \textbf{57.01} & \textbf{40.52} & \underline{50.54} & 34.26 \\ 
        \hline 
        \multirow{1}{*}{ACMM} & 57.27 & 70.85 & 49.19 & 34.02 & 34.90 & 35.63 \\  
        \multirow{1}{*}{ACMP} & 58.41 & 73.85 & 49.06 & 37.44 & 42.48 & 34.57 \\  
        \multirow{1}{*}{ACMMP} & 59.38 & 68.50 & 53.28 & 37.84 & 44.64 & 33.79 \\  
        \multirow{1}{*}{SD-MVS} & 63.31 & 76.63 & 53.78 & 40.18 & 47.37 & 35.53 \\  
        \multirow{1}{*}{HPM-MVS++} & 61.59 & 73.79 & 54.01 & 39.65 & 41.09 & \textbf{40.79} \\  
        \multirow{1}{*}{APD-MVS (base)} & 63.64 & 75.06 & \underline{55.58} & 39.91 & 49.41 & 33.77 \\  
        \hline 
        \multirow{1}{*}{MSP-MVS (ours)} & \textbf{64.48} & \textbf{76.42} & 56.21 & \underline{40.26} & \textbf{51.20} & 33.38 \\  
        \hline
        \end{tabular}%
    }
    \caption{Quantitative results on TNT dataset at given threshold. Our method achieves the highest F$_1$ and completeness.}
    \label{table: TNT results}%
\end{table}%

Moreover, to demonstrate the efficiency of our method, we present a comparative analysis of GPU memory usage and runtime on ETH3D training datasets, as shown in Fig. \ref{fig:mem and time} (c). All experiments are performed on original-resolution images, with the number of images standardized to $10$ across all scenarios for runtime evaluation. Additionally, to ensure fairness, all methods are tested on the same system, with its hardware configuration detailed in the previous section.

Concerning learning-based methods, although IterMVS-LS has the shortest runtime, its memory usage exceeds the maximum capacity of mainstream GPUs. Similarly, other learning-based methods also struggle with excessive memory consumption, making them impractical for large-scale reconstructions. Regarding traditional methods,  while our method requires approximately one-third more time than its baseline, APD-MVS, its memory usage significantly outperforms most methods, including ACMMP, SD-MVS, and HPM-MVS. Thus, our approach can achieve state-of-the-art performance with acceptable runtime and minimal memory consumption, proving its effectiveness and practicality.

\subsection{Ablation Study}
Tab. \ref{table: ablation study} presents the ablation studies of each proposed component. Here we primarily use F$_1$ score for overall comparison.

\subsubsection{Multi-granularity Segmentation Prior} 
We separately remove edge aggregation (w/o. Agr.), edge refinement (w/o. Ref.) and both (w/o. Mul.). w/o. Mul. produces the worst results, highlighting the effectiveness of multi-granularity segmentation prior for edge-confined patch deformation. w/o. Ref. performs better than w/o. Agr., indicating that edge aggregation exhibit a greater impact than edge refinement.

\subsubsection{Attention-Consistent Patch Deformation} 
We respectively exclude the module of adaptive equidistribution (w/o. Eqd.), disassemble-clustering (w/o. Clu.) and both (w/o. Con.). 
w/o. Con. results in the poorest performance, demonstrating the significant impact of attention-consistent patch deformation.
w/o. Clu. slightly outperforms w/o. Eqd., yet both fall short to MSP-MVS, implying that adaptive equidistribution contributes more than disassemble-clustering. 

\subsubsection{Disparity-Sampling Synergistic 3D Optimization} 
We respectively eliminate the variance component (w/o. Var.) and cost component (w/o. Cst.) in objective function, the whole module (w/o. Syn.) and set $\mu$ = 3.
Due to the absence of 3D optimization, w/o. Syn. achieves the poorest results. 
w/o. Var. yields better results than w/o. Cst., suggesting that cost plays a more critical role than variance during cost optimization.
Moreover, $\mu$ = 3 are inferior to $\mu$ = 5 (MSP-MVS), indicating the necessity of cost profile aggregation.

\begin{table}
    \centering
    \renewcommand{\arraystretch}{1.05} 
    \resizebox{\linewidth}{!}{
        \begin{tabular}{c|ccc|ccc}
        \hline
        \multirow{2}{*}{Method} & \multicolumn{3}{c|}{$2cm$} & \multicolumn{3}{c}{$10cm$} \\
        \cline{2-7} & F$_1$ & Comp. & Acc. & F$_1$ & Comp. & Acc. \\   
        \hline  
        \multirow{1}{*}{w/o. Mul.} & 86.53 & 84.32 & \textbf{89.06} & 96.59 & 95.25 & \textbf{97.95} \\   
        \multirow{1}{*}{w/o. Agr.} & 87.07 & 85.88 & 88.48 & 97.01 & 96.38 & 97.64 \\   
        \multirow{1}{*}{w/o. Ref.} & 87.54 & 86.94 & 88.36 & 97.32 & 97.05 & 97.59 \\   
        \hline
        \multirow{1}{*}{w/o. Con.} & 86.61 & 86.49 & 86.91 & 96.77 & 96.88 & 96.66 \\   
        \multirow{1}{*}{w/o. Eqd.} & 87.23 & 87.08 & 87.58 & 97.16 & 97.25 & 97.08 \\   
        \multirow{1}{*}{w/o. Clu.} & 87.29 & 87.16 & 87.62 & 97.21 & 97.29 & 97.14 \\   
        \hline
        \multirow{1}{*}{w/o. Syn.} & 86.94 & 87.29 & 86.8 & 96.9 & 97.33 & 96.47 \\   
        \multirow{1}{*}{w/o. Var.} & 87.17 & 87.41 & 87.15 & 97.07 & 97.46 & 96.68 \\   
        \multirow{1}{*}{w/o. Cst.} & 87.49 & 87.6 & 87.57 & 97.28 & 97.55 & 97.02 \\   
        \multirow{1}{*}{$\mu$ = 3} & 87.73 & 87.53 & 88.13 & 97.42 & 97.59 & 97.25 \\   
        \hline
        \multirow{1}{*}{MSP-MVS} & \textbf{88.09} & \textbf{87.87} & 88.51 & \textbf{97.70} & \textbf{97.74} & 97.67 \\  
        \hline
        \end{tabular}%
    }
    \caption{Quantitative results of the ablation studies on ETH3D benchmark to validate each proposed component.}
    \label{table: ablation study}%
\end{table}%

\section{Conclusion}
In this paper, we introduce MSP-MVS, a method leveraging multi-granularity segmentation prior for edge-confined patch deformation. We first aggregate and refine multi-granularity depth edges via Semantic-SAM as prior to guide patch deformation within homogeneous areas.
Moreover, we implement adaptive equidistribution for sector division and disassemble-clustering for anchor clustering to enable patch deformation with attention consistency. Additionally, we propose disparity-sampling synergistic 3D optimization to help deformed patches identify global-minimum matching costs.
Results on ETH3D and Tanks \& Temples benchmarks validate the state-of-the-art performance of our method.

\section{Acknowledgements}
This work was supported in part by the National Natural Science Foundation of China under Grant 62172392, in part by the Strategic Priority Research Program of the Chinese Academy of Sciences under Grant No. XDA0450203, in part by the Innovation Program of Chinese Academy of Agricultural Sciences (Grant No. CAAS-CSSAE-202401 and CAAS-ASTIP-2024-AII), and in part by the Beijing Smart Agriculture Innovation Consortium Project (Grant No. BAIC10-2024).

\newpage

\section{Supplementary Material}
In this supplementary material, we significantly expand upon the related work discussed in the main paper to provide a comprehensive view of the research landscape surrounding MSP-MVS. We structure this discussion into two main sections. The first section, "Expanded Related Work," situates our method within the direct lineage of 3D computer vision, from classical geometric principles to modern neural representations. The second section, "Broader Context and Future Directions," elevates the perspective, connecting MSP-MVS to the paradigm-shifting developments in foundation models and cross-disciplinary AI research, thereby outlining a vision for the future of 3D perception.

\subsection{More Details on Implementation}
Our method is implemented on a machine with an Intel(R) Xeon(R) Silver 4210 CPU and eight NVIDIA GeForce RTX 3090 GPUs. Experiments are performed on original images in both ETH3D dataset and TNT dataset.
In cost calculation, we adopt the matching strategy of every other row and column. 
We take APD-MVS \cite{APD-MVS} as our baseline.

\section{Results on ETH3D and TNT dataset}
\begin{figure*}
    \centering
    \includegraphics[width=\linewidth]{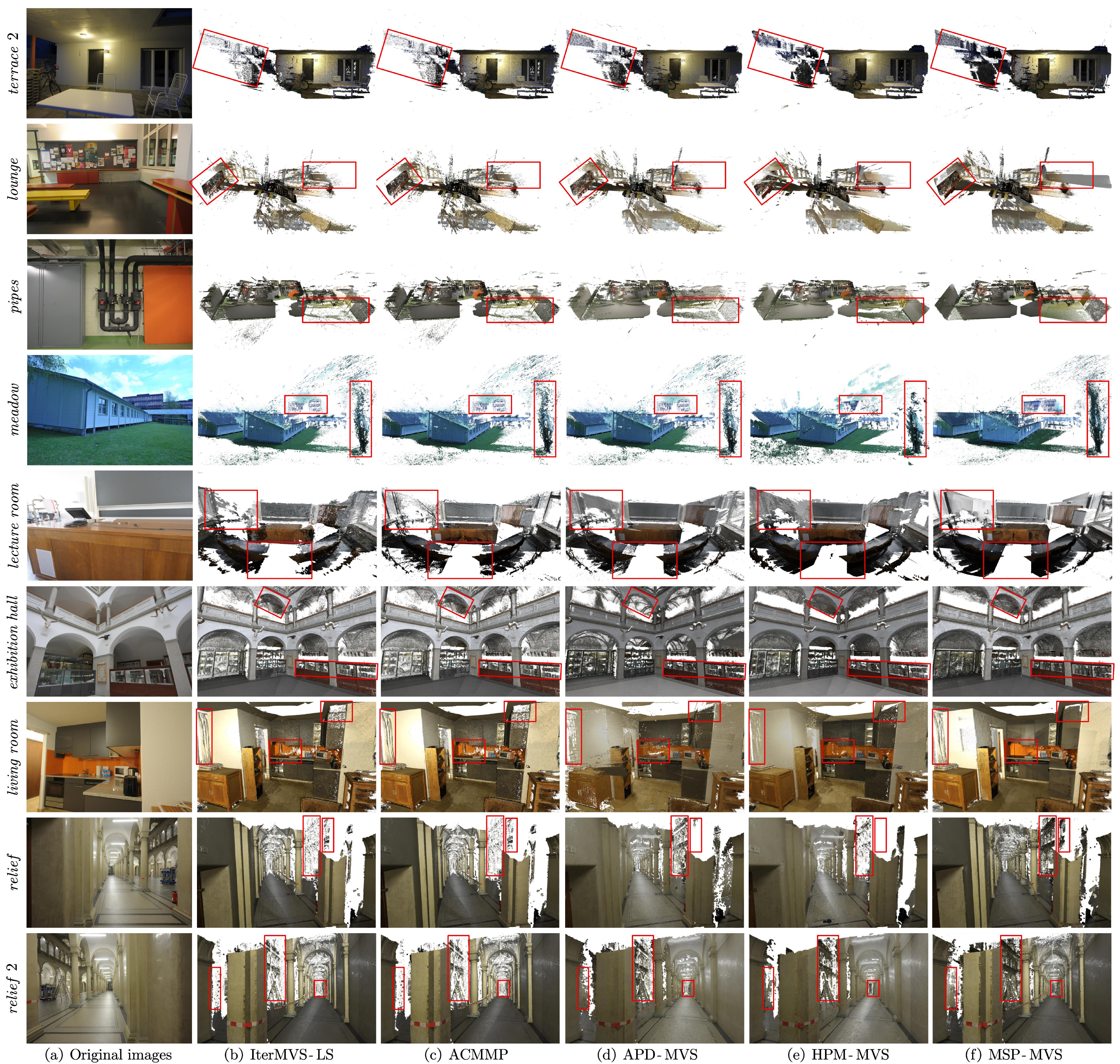}
    \caption{An illustration of the qualitative results on partial scenes of ETH3D datasets (\emph{terrace 2}, \emph{lounge}, \emph{pipes}, \emph{meadow}, \emph{lecture room}, \emph{exhibition hall}, \emph{living room}, \emph{relief} and \emph{relief 2}). It is obvious that our methods outperform others, especially in large textureless areas.}
    \label{fig:eth3d_supp}
\end{figure*}

\begin{figure*}
    \centering
    \includegraphics[width=\linewidth]{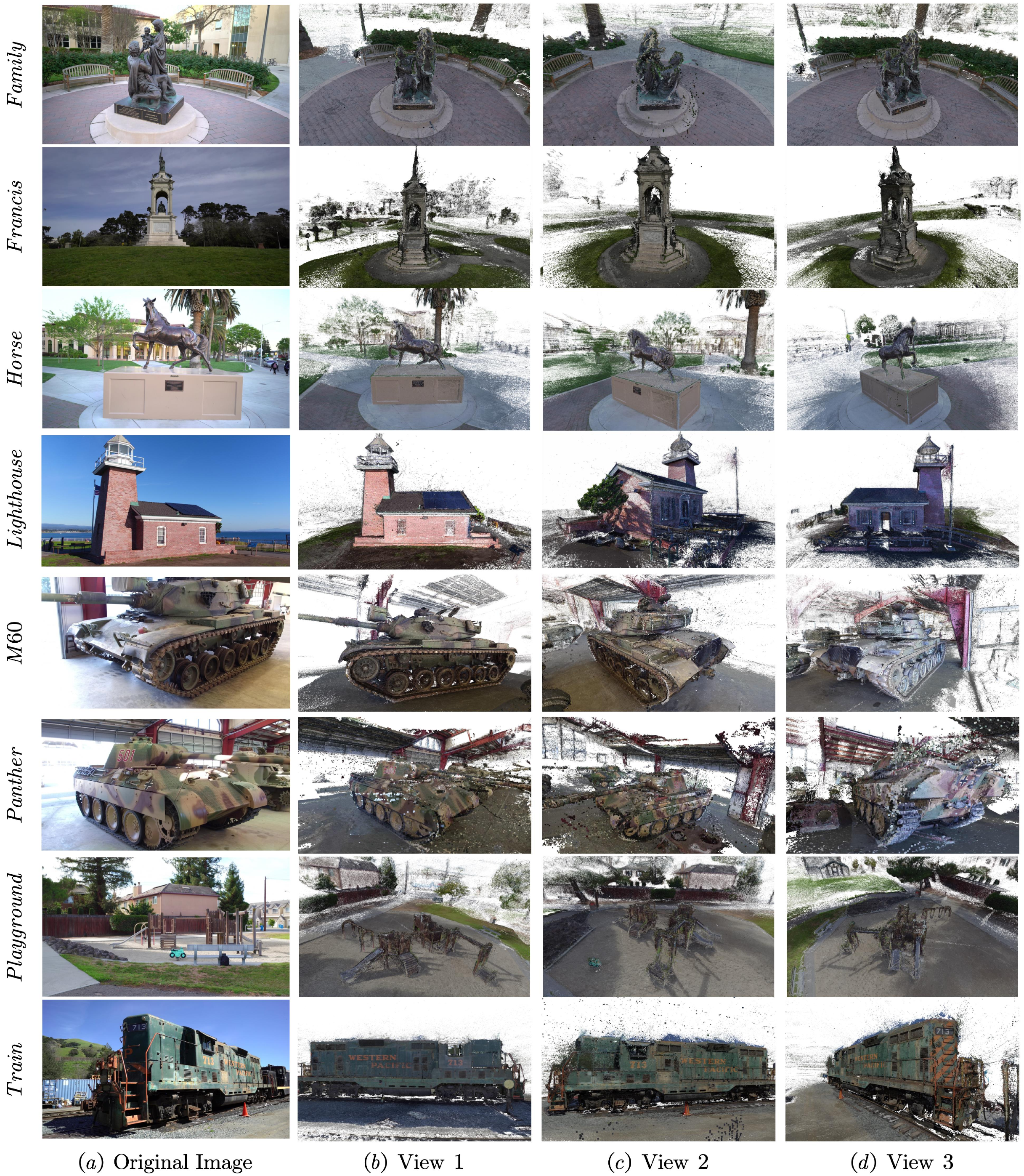}
    \caption{Reconstructed point clouds results on intermediate set of TNT datasets.}
    \label{fig:tnt_intermediate}
\end{figure*}

\begin{figure*}
    \centering
    \includegraphics[width=\linewidth]{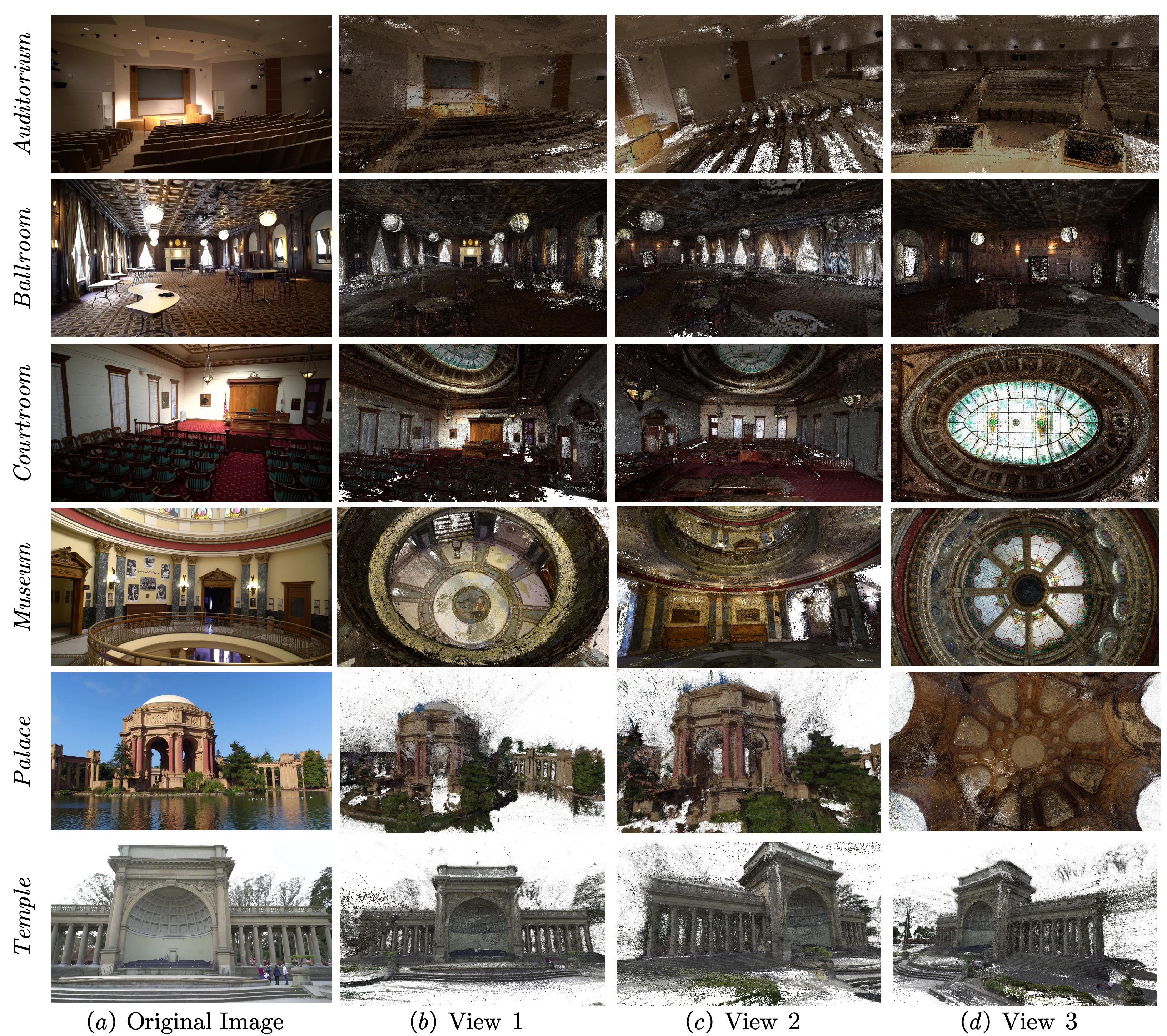}
    \caption{Reconstructed point clouds results on advanced set of TNT datasets.}
    \label{fig:tnt_advanced}
\end{figure*}

Fig. \ref{fig:eth3d_supp} demonstrates qualitative comparisons among various methods on partial scenarios of the ETH3D datasets. Similarly, Fig. \ref{fig:tnt_intermediate} and Fig. \ref{fig:tnt_advanced} illustrate the qualitative results in the intermediate and advanced sets of the Tanks \& Temples datasets, respectively. As can be seen, our method reconstructs the most complete and realistic point clouds than other state-of-the-art traditional MVS methods, especially in recovering large textureless areas such as walls and floors, without introducing detail distortion, evidencing its strong effectiveness and robust generalization ability.

\section{Analyses and Discussions}
\par
\noindent\textbf{Q1: Compared to fixed interval sampling or global sampling, what is the success rate or accuracy of the adaptive sampling strategy in Synergistic 3D Optimization?}
\par
\noindent\textbf{A1:} 
As shown in Tab. \ref{table: ablation study}, we respectively perform the fixed interval sampling strategy (Fix.), the global sampling strategy (Glo.) and the adaptive sampling strategy (Apt.) on deformed patches of ETH3D training dataset for comparison. 
Evidently, the F$_1$ scores of Apt. significantly outperform Fix., indicating the effectiveness of adaptive sampling strategy. 
While Glo. achieves a slightly better F$_1$ score than Apt., it requires over 5 times the runtime compared to our method, indicating that our adaptive sampling strategy can achieve better trade-offs between time efficiency and performance.

\begin{figure}
\centering
\includegraphics[width=\linewidth]{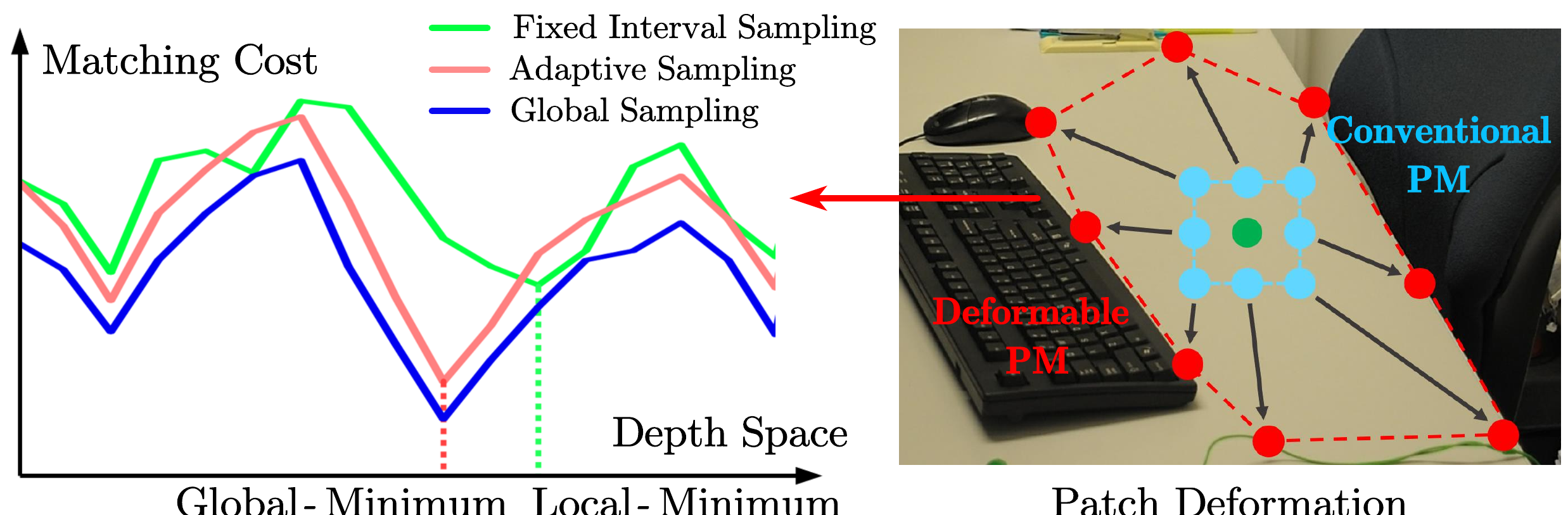}
\caption{Comparative 2D cost profiles of different sampling strategies. The green, blue, red line in left respectively denote fixed interval, global and our adaptive sampling strategy employed in red deformable PM anchors of right graph.
}
\vspace{-0.1in}
\label{fig: cost profile}
\end{figure}

Moreover, as shown in Fig. \ref{fig: cost profile}, we present different 2D cost profiles of deformed patches employed by different sampling strategies for intuitive comparison. Obviously, the fixed interval sampling strategy (i.e. green line) inevitably causes cost to fall into a local-minimum. Additionally, while the cost profile of global sampling strategy (i.e. blue line) achieves a lower overall cost, both the global and adaptive sampling strategies (i.e. red line) can approximate the global-minimum matching cost. Consequently, our adaptive sampling strategies can identify the global minimum cost with less time, demonstrating its high efficiency.

\par
\noindent\textbf{Q2: There are already MVS methods based on semantic segmentation like SD-MVS \cite{SD-MVS} and Multi-View SAM \cite{Shvets2024}, what's our advantage?}
\par
\noindent\textbf{A2:} 
While existing methods like SD-MVS and Multi-View SAM utilize semantic segmentation, they typically leverage semantic information at a single granularity, which often fails to capture comprehensive depth edges in complex scenarios. In contrast, our approach can integrate and further refine semantic information across multiple granularities, allowing us to aggregate more robust depth edges. 
For instance, when reconstructing a book, text and images on its cover might be mistakenly identified as depth edges by SD-MVS. However, our MSP-MVS can effectively exclude such distractions, treating them together as homogeneous areas for patch deformation, thus improving matching accuracy.

\par
\noindent\textbf{Q3: In the Disassemble-Clustering strategy, what function does the disassemble operation serve?}
\par
\noindent\textbf{A3:} 
The disassemble operation in disassemble-clustering strategy serves to maintain the overall patch size while allowing our method to select more reliable anchors. By quadrupling the number of anchors and disassembling the original patch into multiple smaller sub-patches, the disassemble operation guarantees the deformed patches covering more reliable pixels without increasing computational complexity.

\begin{table}
    \centering
    \renewcommand{\arraystretch}{1.01} 
    \resizebox{\linewidth}{!}{
        \begin{tabular}{c|ccc|ccc}
        \hline
        \multirow{2}{*}{Method} & \multicolumn{3}{c|}{$2cm$} & \multicolumn{3}{c}{$10cm$} \\
        \cline{2-7} & F$_1$ & Comp. & Acc. & F$_1$ & Comp. & Acc. \\   
        \hline  
        \multirow{1}{*}{Fix.} & 86.94 & 87.29 & 86.8 & 96.9 & 97.33 & 96.47 \\    
        \multirow{1}{*}{Glo.} & \textbf{88.45} & \textbf{88.32} & \textbf{88.78} & \textbf{97.94} & \textbf{98.03} & \textbf{97.85} \\   
        \hline
        \multirow{1}{*}{Apt.} & 88.09 & 87.87 & 88.51 & 97.70 & 97.74 & 97.67 \\  
        \hline
        \end{tabular}%
    }
    \caption{Quantitative ablation studies on ETH3D dataset of different sampling strategies employed in deformed patches.}
    \vspace{-0.1in}
    \label{table: ablation study}%
\end{table}%

\par
\noindent\textbf{Q4: How exactly does the conditional random field algorithm achieve edge refinement?}
\par
\noindent\textbf{A4:} 
According to \cite{CRF2001}, the conditional random field can leverage inter-pixel correlation to refine label distribution, which is achieved by modeling the predictions within a graphical model that illustrates the dependencies among them. 

Specifically, in the main text, we have already defined the unary potential $\psi_u\left(l_i\right)$ and the pairwise potential $\psi_p\left(l_i, l_j\right)$. We then obtain the overall energy function $\Psi(L)$ by aggregating $\psi_u\left(l_i\right)$ and $\psi_p\left(l_i, l_j\right)$ among all pixels, calculated by:
\begin{equation}
\Psi(L)=\sum_i^K \psi_u\left(l_i\right)+\sum_{i, j}^K \psi_p\left(l_i, l_j\right).
\end{equation}
By minimizing $\Psi(L)$ CRF effectively achieves edge refinement.
To accomplish the minimization process, an observation sequence $X$ and a corresponding label sequence $Y$ are defined to express the conditional probability of the model:
\begin{equation}
P(Y \mid X)=\frac{1}{Z(X)} \exp \left(\sum_k \lambda_k \Psi_k\left(Y, X\right)\right),
\end{equation}
where $X = \{x_1, x_2, ..., x_n\}$ and $Y = \{y_1, y_2, ..., y_n\}$ respectively represents all pixels and their corresponding labels, $\Psi_k\left(Y, X\right)$ denotes the feature function, each feature function corresponds to a weight $\lambda_k$, and $Z(X)$ is the normalization factor to ensure a valid probability distribution, computed by summing over all possible label configurations $Y$. Subsequently, to maximize the prediction accuracy $\mathcal{L}(\Lambda)$, the algorithm employs the maximization of the log-likelihood function to optimize the parameters $\lambda_k$. Mathematically, $\mathcal{L}(\Lambda)$ is calculated by:
\begin{equation}
\mathcal{L}(\Lambda)=\sum_{(X, Y)} \log P(Y \mid X ; \Lambda)-\frac{1}{2 \eta^2}\|\Lambda\|^2,
\end{equation}
where $\Lambda = \{\lambda_1, \lambda_2, ..., \lambda_k\}$ represents the set of all feature function weights, $\eta$ is a regularization term to prevent overfitting.
The maximization of $\mathcal{L}(\Lambda)$ can be achieved through the Quasi-Newton method \cite{CRF2012}.

\subsection{Expanded Related Work}
This section contextualizes MSP-MVS within the broader evolution of Multi-View Stereo (MVS) and 3D scene representation. We begin by delving into the advancements in patch-based MVS, transition to the rise of learning-based 3D representations, discuss the critical role of semantic and geometric priors, and conclude with the diverse applications of these technologies, illustrating the impact and relevance of robust 3D reconstruction.

\subsubsection{Advancements in Patch-based and Hybrid MVS}

Traditional MVS methods have long relied on the PatchMatch algorithm~\citep{PM}, which iteratively refines depth estimates by propagating and matching local patches. However, these methods often falter in texture-poor regions where patch-based matching becomes ambiguous. To address this, hybrid approaches have emerged, augmenting geometric principles with more robust strategies. Our work, MSP-MVS, falls into this category by introducing multi-granularity segmentation priors to guide patch deformation. This line of research, focusing on enhancing reconstruction in challenging areas, has seen significant recent activity. For instance, similar to our use of semantic edges, other works have proposed using dual-level precision edges to guide MVS and ensure accurate planarization~\citep{chen2025dual}. The core idea of leveraging segmentation for robust reconstruction is further explored in methods that are explicitly segmentation-driven, incorporating sophisticated mechanisms like depth restoration and occlusion handling to improve results~\citep{yuan2025sed}. An earlier precursor to these ideas can be seen in textureless-aware segmentation and correlative refinement guided MVS, which also highlights the value of semantic understanding in traditional MVS pipelines. These methods collectively demonstrate a clear trend: the fusion of high-level semantic guidance with low-level geometric matching is crucial for achieving state-of-the-art performance.

\subsubsection{The Rise of Learning-based 3D Representations}

While hybrid methods enhance traditional pipelines, the field has been revolutionized by end-to-end learning-based approaches. This paradigm shift began with MVS networks that replaced handcrafted cost functions with learned 3D cost volumes. Recent advancements in this domain have focused on improving network architectures, for instance, by leveraging monocular priors to guide the multi-view process~\citep{monomvsnet} or by incorporating powerful backbones like Recurrent Regularization Transformers~\citep{rrt-mvs} and State Space Models~\citep{mvsmamba} to better aggregate information across views.

Concurrently, a new frontier opened with the development of implicit and explicit neural representations for scenes. Neural Radiance Fields (NeRF) offered a way to synthesize photorealistic novel views from a continuous volumetric function, and community efforts have produced comprehensive datasets to benchmark such NeRF-based 3D reconstruction techniques~\citep{yan2023nerfbk}. More recently, 3D Gaussian Splatting (GS) has gained immense popularity due to its ability to achieve real-time rendering with high fidelity. Research in this direction is rapidly expanding, with a focus on several key areas. These include developing lightweight and compact models for large-scale dynamic 4D scenes through context-aware modeling~\citep{liu2025light4gs, liu2025light4gs}, enabling real-time video representation at high frame rates through deformable 2D Gaussians~\citep{liu2025d2gv}, and creating intuitive tools for controllable 3D scene editing~\citep{yan20243dsceneeditor}. Furthermore, efforts are being made to enhance the structural integrity of GS reconstructions by leveraging topological concepts like persistent homology~\citep{shen2025topology}, and to unify appearance codes with bilateral grids for complex driving scene reconstruction~\citep{wang2025unifying}. A critical challenge is extending these representations to encompass not just static scenes but the world's dynamics, leading to the development of world models with self-supervised 3D labels for robotics applications~\citep{yan2025renderworld}. Bridging the gap between these new representations and classic computer vision tasks, recent work has also focused on performing object segmentation directly on the Gaussian representation, guided by gradient information to enhance boundary precision~\citep{li2024gradiseg}, and enabling spatio-temporal decoupling for real-time dynamic scene rendering~\citep{li2025stdr}.

\subsubsection{Semantic and Geometric Priors in 3D Perception}

The core innovation of MSP-MVS is the use of multi-granularity priors from Semantic-SAM. This highlights a broader principle: leveraging structured information is key to solving ill-posed perception problems. This principle is widely applied across various domains. In the context of segmentation, for instance, researchers have explored scaling up autoregressive pretraining for complex structures like neurons~\citep{chen2025tokenunify}, using dual-form complementary masking for domain-adaptive image segmentation~\citep{wang2025masktwins}, and employing multi-agent reinforcement learning for self-supervised segmentation~\citep{chen2023self}.

Beyond segmentation, graph-based models offer a powerful way to represent structural relationships. Techniques such as multi-scale graph learning for challenging tasks like anti-sparse downscaling~\citep{fan2025multi} demonstrate the versatility of graph structures. In the realm of graph autoencoders, revisiting masking strategies from a robustness perspective~\citep{song2025equipping} and developing self-purified models to release robust expression power~\citep{song2025spmgae} have pushed the boundaries of representation learning. For composed image and video retrieval, a deep understanding of fine-grained semantics is paramount. This has led to methods that explicitly parse modification semantics~\citep{FineCIR}, perform visual-semantic decomposition and partial alignment~\citep{EmDepart}, disentangle representations with complementarity guidance~\citep{PAIR}, utilize hierarchical uncertainty-aware disambiguation~\citep{HUD}, revise focus shifts based on segmentation~\citep{OFFSET}, and mine entity and modification relations~\citep{ENCODER}. These diverse approaches underscore the universal importance of leveraging semantic and structural priors, a principle that lies at the heart of our MSP-MVS.

\subsubsection{Applications of 3D Reconstruction and Perception}

The ultimate goal of advancing 3D perception is to enable robust real-world applications. Autonomous Driving is a primary beneficiary. Accurate 3D reconstruction and perception are essential for building online HD maps~\citep{zhang2025mapexpert}, predicting trajectories from a cross-view perspective~\citep{song2023xvtp3d}, and estimating depth under adverse weather conditions through curriculum contrastive learning~\citep{Wang_2024}. This same line of inquiry has also explored the synergy between contrastive learning and diffusion models for even more robust depth estimation, demonstrating the rapid evolution of perception techniques in this area~\citep{Wang_2024_2}. Recent work even explores transforming image editors into dense geometry estimators~\citep{wang2025editor} and leveraging adversarial transfer for end-to-end driving models~\citep{zhangdrive}. Simplified yet effective models with distinct experts and implicit interactions are also being developed to tackle the complexity of end-to-end driving~\citep{zhangaddi}.

In Remote Sensing, multi-view graph clustering with dual relation optimization~\citep{MDRO}, structure-adaptive multi-view graph clustering~\citep{SAMVGC}, and sampling-enhanced contrastive clustering with long-short range information mining~\citep{SEC-LSRM} are all techniques developed to handle the unique challenges of satellite and aerial imagery. The development of interactive agents for comprehensive change interpretation and analysis~\citep{liu2024changeAgent} and comprehensive surveys on spatiotemporal vision-language models for remote sensing~\citep{liu2025RSTVLM_survey} further highlight the rapid progress in this field. Diffusion models are also being used for controllable image generation~\citep{tang2024crs} and to enhance object detection through data generation~\citep{tang2025aerogen}.

Finally, in Robotics and Human-Object Interaction (HOI), discovering syntactic interaction clues~\citep{luo2024discovering} and employing synergistic prompting learning~\citep{luo2025synergistic} are crucial for enabling machines to understand and interact with their environment in a human-like manner.

In summary, this expanded review situates MSP-MVS at the confluence of several critical research streams. By building upon the robust foundations of patch-based MVS while integrating high-level semantic priors, our work addresses long-standing challenges in textureless reconstruction. It stands as a complementary approach to the concurrent rise of learning-based representations like NeRF and Gaussian Splatting, offering a pathway that prioritizes geometric accuracy and computational efficiency. As demonstrated by the diverse and demanding applications in autonomous driving and remote sensing, the need for reliable and precise 3D perception remains paramount. MSP-MVS contributes a significant step forward in this direction, proving that the principles of guided geometric matching remain highly relevant in the modern 3D vision landscape.

\subsection{Broader Context and Future Directions: The Role of Foundation Models and Cross-Disciplinary Insights}
Having situated MSP-MVS within the context of 3D vision, we now broaden our perspective to discuss its connections to the wider trends in artificial intelligence. This section explores how the rise of foundation models, advances in general learning paradigms, and insights from specialized, cross-disciplinary domains are poised to shape the future of MVS and 3D perception at large.

\subsubsection{The Era of Vision-Language Models (VLMs): Towards Language-Guided 3D Systems}

The rise of Large Language Models (LLMs) and Vision-Language Models (VLMs) is reshaping computer vision. The ability of these models to reason, follow instructions, and process multi-modal information offers exciting possibilities for MVS. Our use of Semantic-SAM is a step in this direction, but future systems could be guided by more complex, natural language commands. Research into VLM capabilities is vast, spanning from enhancing LLMs' confidence on edited facts~\citep{bi2024decoding} and aligning them for context-faithfulness~\citep{bi2024context} to controlling knowledge reliance through fine-grained mechanisms~\citep{bi2025parameters} and promoting exploration to improve multi-domain reasoning~\citep{bi2025reward}.

VLMs are being applied to a wide array of tasks that hint at their potential for 3D. This includes versatile advertising poster layout generation~\citep{anonymous2025anylayout}, explainable visual question answering via diffusion-based chain-of-thought~\citep{lu2024explainable}, and challenging VLMs' spatial intelligence through complex reasoning tasks~\citep{song2025siri}. For specific applications, VLMs are being used for document-based zero-shot learning by progressively optimizing prompts~\citep{ProAPO} and leveraging multi-attribute document supervision~\citep{MADS}. Instruction-based learning is also proving effective for complex tasks like HOI detection~\citep{luoinstructhoi}. The synergy between retrieval mechanisms and preference optimization is being used to better align VLMs~\citep{xing2025re}, and hierarchical cross-modal alignment is being developed for decoupled multimodal representation learning~\citep{qian2025decalign}. The most direct link to 3D is seen in generative text-guided 3D vision-language pretraining for medical image segmentation~\citep{chen2025gtgm}, which paves the way for future language-guided 3D reconstruction systems.

\subsubsection{Advanced Learning Paradigms and Their Implications for MVS}

Beyond VLMs, advances in general machine learning paradigms offer promising directions for MVS. Data-centric AI emphasizes the quality of training data. This includes learning to refine pre-training data at scale from expert-guided programs~\citep{bi2025refinex} and developing robust digital watermarking frameworks resistant to extreme cropping, scaling~\citep{sunultra}, and non-differentiable distortions~\citep{sun2025end} to ensure data integrity.

In terms of model architectures and training, new models like the diffusion Mamba for end-to-end multi-modal tasks~\citep{lu2025end} and lightweight model transition methods to improve LLM training via weight scaling~\citep{li2025wiscalightweightmodeltransition} could lead to more efficient and powerful MVS networks. Furthermore, in the realm of graph-based learning, research into elevating the resilience of graph prompt tuning against adversarial attacks~\citep{song2025gpromptshield} is critical for building robust systems.

Finally, the concept of intelligent agents is gaining traction. This includes the development of world cognition agents enabling adaptive human-AI symbiosis in industrial settings~\citep{liu2025mr} and comprehensive surveys on pure Vision Language Action (VLA) models~\citep{zhang2025pure}. Such agents could one day perform 3D reconstruction as a tool to achieve a broader goal. For autonomous driving, VLA models are being developed to incentivize reasoning and self-reflection~\citep{yuan2025autodrive}, while physical autoregressive models are being created for robotic manipulation without action pretraining~\citep{song2025physical}. Reinforcing open-vocabulary action recognition with tools is another step towards more capable and generalist agents~\citep{yuan2025video}.

\subsubsection{Cross-Disciplinary Insights from Specialized Domains}

Often, solutions developed for niche problems can inspire general-purpose algorithms. Medical Imaging, a field with unique challenges like extreme low-texture regions and noisy labels, is a fertile ground for such insights. Techniques developed for medical image segmentation, such as leveraging the frequency domain~\citep{han2025frequency}, estimating region uncertainty to handle noisy labels~\citep{han2025region}, and using adaptive label correction~\citep{qian2025adaptive}, could be adapted to improve the robustness of MVS in ambiguous scenarios. Similarly, curriculum learning frameworks for imbalanced multimodal diagnosis~\citep{qian2025dyncim, han2025climd} could inspire new training strategies for MVS on unbalanced datasets. For cryo-electron tomography, a field with notoriously low signal-to-noise ratio, methods for self-supervised volumetric image restoration~\citep{Yang_2021_ICCV}, denoising based on noise modeling and transfer~\citep{n2tc, nmsg}, and improved denoising with simulation-aware pretraining~\citep{cmb20240513} are highly relevant to MVS's own struggle with noise and texturelessness. Furthermore, the use of sparse and hierarchical transformers for survival analysis on whole slide images~\citep{10226279} and the fusion of multi-scale heterogeneous pathology foundation models~\citep{yang2025fusionmultiscaleheterogeneouspathology} offer architectural ideas for efficient multi-scale feature aggregation. The development of evolutionary medical prompt optimization~\citep{chen2025empower} showcases a powerful method for automatically tailoring inputs to large models, a technique detailed in related publications~\citep{11205280} that could be applied to guide future 3D generative systems.

Insights also come from other industrial applications. For example, efficient hybrid-supervised symmetric stereo matching networks~\citep{Zhang2023EHSSAE} and effective classification methods for steel surface defects based on salient local features~\citep{hao2024ssdc} provide battle-tested solutions for high-precision industrial vision tasks. Even a cloud framework designed for autonomous driving path planning~\citep{10858860} points towards the future of MVS as a service in large-scale, distributed systems. By drawing inspiration from these diverse fields, the MVS community can continue to build more robust, efficient, and versatile 3D perception systems. Finally, the principles of contrastive learning, enhanced with diffusion models, are being used for robust depth estimation~\citep{Wang_2024}, and harnessing diffusion priors for self-supervised depth estimation~\citep{wang2025jasmine} shows another promising direction for future MVS research.

In conclusion, by viewing MSP-MVS through the lens of broader AI advancements, we can appreciate its role not merely as an algorithmic improvement, but as a harbinger of future 3D perception systems. Its use of a foundation model for semantic guidance is indicative of a larger trend towards language-guided, context-aware intelligence. The ongoing revolutions in advanced learning paradigms and the surprising yet potent insights from cross-disciplinary fields like medical imaging provide a rich tapestry of ideas for the next generation of MVS. The future of 3D reconstruction will likely involve a deeper symbiosis with large models, more sophisticated data-centric strategies, and the integration of reasoning and action capabilities, transforming MVS from a passive reconstruction tool into an active perception module within larger intelligent agents. Our work represents a concrete step on this exciting path.

\bibliography{aaai25}

\end{document}